\documentclass[letterpaper, 10 pt, conference]{ieeeconf}  % Comment this line out if you need a4paper

\IEEEoverridecommandlockouts                              % This command is only needed if 
\usepackage{graphics} % for pdf, bitmapped graphics files
\usepackage{subfigure}
\usepackage{amsmath,amssymb}
\usepackage{multirow}
\usepackage[pdftex]{graphicx}
\usepackage{eqparbox}
\usepackage{float}
\usepackage{booktabs}
\usepackage{tabu}
\usepackage{color, colortbl}
\usepackage{bbding} %首先在导言区调用bbding包
\usepackage{pifont}

\usepackage{booktabs, cellspace, multirow, tabularx}
    \newcolumntype{R}{>{\raggedright\arraybackslash}X}
    \addparagraphcolumntypes{R}
\usepackage{ragged2e}
\usepackage{stackengine} 
\newcommand\oast{\stackMath\mathbin{\stackinset{c}{0ex}{c}{0ex}{\ast}{\bigcirc}}}

\usepackage[table]{xcolor}
\usepackage[export]{adjustbox}
\usepackage{amsmath}
\usepackage[colorlinks,allcolors=blue]{hyperref}

\title{\LARGE \bf
Toward Accurate Camera-based 3D Object Detection via Cascade Depth Estimation and Calibration
}
\author{Chaoqun Wang$^{1}$,
Yiran Qin$^{1}$, \thanks{$^{1}$ School of Data Science, Shenzhen Research Institute of Big Data, The Chinese University of Hong Kong, Shenzhen (CUHK-Shenzhen), China. \{chaoqunwang@link., yiranqin@link., ruimaozhang@\}cuhk.edu.cn}
Zijian Kang$^{2}$, \thanks{$^{2}$ NIO. \{zijian.kang, ningning.ma\}@nio.com}
Ningning Ma$^{2}$, 
and
Ruimao Zhang$^{1\dag}$\thanks{\dag Corresponding author.}\\
}

\begin{document}

\maketitle
\thispagestyle{empty}
\pagestyle{empty}

%%%%%%%%%%%%%%%%%%%%%%%%%%%%%%%%%%%%%%%%%%%%%%%%%%%%%%%%%%%%%%%%%%%%%%%%%%%%%%%%
\begin{abstract}
Recent camera-based  3D object detection is limited by the precision of transforming from image to 3D feature spaces, as well as the accuracy of object localization within the 3D space. This paper aims to address such a fundamental problem of camera-based 3D object detection: How to effectively learn depth information for accurate feature lifting and object localization. Different from previous methods which directly predict depth distributions by using a supervised estimation model, we propose a cascade framework consisting of two depth-aware learning paradigms. First, a depth estimation (DE) scheme leverages relative depth information to realize the effective feature lifting from 2D to 3D spaces. Furthermore, a depth calibration (DC) scheme introduces depth reconstruction to further adjust the 3D object localization perturbation along the depth axis. In practice, the DE is explicitly realized by using both the absolute and relative depth optimization loss to promote the precision of depth prediction, while the capability of DC is implicitly embedded into the detection Transformer through a depth denoising mechanism in the training phase. The entire model training is accomplished through an end-to-end manner. We propose a baseline detector and evaluate the effectiveness of our proposal with +2.2\%/+2.7\% NDS/mAP improvements on NuScenes benchmark, and gain a comparable performance with 55.9\%/45.7\% NDS/mAP. Furthermore, we conduct extensive experiments to demonstrate its generality based on various detectors with about +2\% NDS improvements. 

\end{abstract}

%%%%%%%%%%%%%%%%%%%%%%%%%%%%%%%%%%%%%%%%%%%%%%%%%%%%%%%%%%%%%%%%%%%%%%%%%%%%%%%%
\section{INTRODUCTION}
Object detection in 3D space is pivotal for numerous real-world applications, including robotics and autonomous driving systems. Recently, camera-based 3D detectors have attracted increasing attention both in academia and the industry, due to various benefits they offer, such as cost-effectiveness, the long perception range, and the capability to recognize vision-only attributes ( \textit{e.g.,} traffic lights and stop signs ). However, when compared to LiDAR-based methods, camera-based systems inherently suffer from a limitation tied to 2D images: the absence of depth information, which is crucial for accurate 3D detection.

\begin{figure}[thpb]
  \centering
  \includegraphics[width=0.45\textwidth]{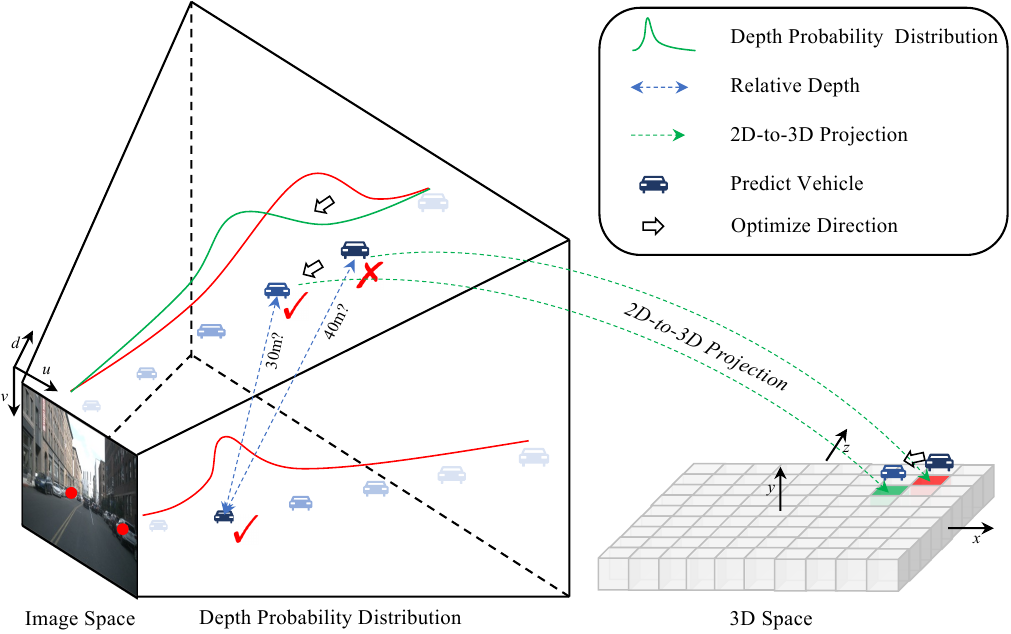}
  \vspace{-3mm}
  \caption{The \textcolor{red}{red} curves indicate the predicted depth probability distribution of objects in \textcolor{red}{red} dots. By supervising relative depth in the DE scheme, we can optimize the distribution~(from \textcolor{red}{red} to \textcolor{green}{green} curve), which is more accurate. In the DC scheme, we generate noised anchors~(\textcolor{red}{red} cube) from the ground truth box~(\textcolor{green}{green} cube) in 3D space. By reconstructing them, the detector obtains depth calibration capability.}
  \label{fig:motivation}
  \vspace{-5mm}
\end{figure}

The literature presents numerous methods aimed at addressing this challenge by extracting depth information. One branch involves directly predicting 3D object boxes from 2D input~\cite{wang2021detr3d, liu2022petr, li2022bevformer}. In these methods, depth information is implicitly learned by optimizing the 3D box location and sizes, supervised by the provided ground truth.
For more precise depth information, another branch of methods~\cite{li2023bevdepth, huang2021bevdet, li2023bevstereo} explicitly introduces the depth estimation module, which is trained by the supervision of depth maps sourced from LiDAR inputs.
Although significant progress has been made, limited by the natural drawback of camera inputs, depth information prediction remains a challenge due to the following two difficulties.
\begin{itemize}
\item Learning the depth usually relies on simple pixel-level supervision and lacks constraints based on the structural context of the scene, leading to high prediction errors.
\item Inaccurate depth information can further exacerbate disturbances in the object localization process, leading to a further decline in the accuracy of 3D detection.
\end{itemize}
To address these coherent issues, we propose a cascade framework leveraging two depth-aware learning paradigms to effectively predict depth information for accurate camera-based 3D object detection.
We first introduce a novel depth estimation (DE) scheme, which defines depth relationships through the context of the scene, and uses it as a supervisory signal to train a more robust depth estimator.
As illustrated on the left of Fig.~\ref{fig:motivation}, given two objects (denoted by \textcolor{red}{red} dot) in the image, when directly using depth information for supervision, the error rate of the trained depth estimator often gradually increases with the growing distance of the objects from the camera. However, by introducing the relative depth between objects at different distances as constraints during training, the aforementioned issue can be effectively mitigated (\textit{e.g.,} the depth probability distribution is corrected from the \textcolor{red}{red} curve to the \textcolor{green}{green} one).

On the other hand, we further introduce a depth calibration (DC) scheme to prevent the decline in object localization accuracy caused by depth estimation errors. During the training process, we introduce a depth denoising mechanism, embedding this capability within the Transformer model.
As shown on the right of Fig.~\ref{fig:motivation}, inaccurate depth estimation in images can lead to objects shifting along the depth axis in 3D space. Our proposed DC scheme simulates this object perturbation, generating noised anchors from the ground truth box. By reconstructing the actual boxes from these noised ones, the detector inherently acquires a depth calibration capability.

In practice, our framework integrates the aforementioned auxiliary supervisions, and training in an end-to-end manner to enhance detection performance without added inference costs. 
In addition, DE and DC schemes can be used independently. Integrating either one into existing camera-based detectors consistently results in a stable performance improvement.
The main contributions can be summarized as follows. 
\begin{itemize}
    \item We present a cascade framework for camera-based 3D object detection, which gains depth prediction accuracy at both the feature lifting and object localization phases. It provides an effective way to mitigate the accuracy decline caused by the cumulative errors of depth prediction in the pipeline.
    \item We introduce the novel depth estimation (DE) and depth calibration (DC) schemes into our framework. These methods leverage pre-defined depth-aware content within scenes for model training, offering a cost-effective approach that significantly enhances the robustness of the detectors.
    \item We have implemented a detector and achieved substantial improvements, demonstrating the efficacy of our proposed framework. Moreover, we have carried out extensive experiments using a variety of open-sourced, state-of-the-art detectors to showcase the broad applicability of both DE and DC schemes.
\end{itemize}

\section{RELATED WORKS}
\textbf{Camera-based 3D Object Detection} requires predicting 3D boxes only from camera data, in which depth estimation is the most challenging since it is an ill-posed problem. One branch of previous works lies in explicitly using depth information~\cite{li2023bevdepth, huang2021bevdet, li2023bevstereo, reading2021categorical, shi2021geometric, li2019gs3d, li2019stereo}. Some work~\cite{li2023bevdepth, huang2021bevdet, li2023bevstereo, reading2021categorical} lift the features from Field-Of-View~(FOV) to Bird-Eye-View~(BEV) plane, via the prediction depth and camera intrinsics/extrinsic matrix, or mimic the pseudo-LiDAR by projecting the depth map into 3D space,
and feed the BEV/pseudo-LiDAR features into detection head to predict 3D box. To gain better performance, BEVDepth~\cite{li2023bevdepth} utilizes the corresponding LiDAR data to generate depth ground truth and supervise the accurate depth estimation. Besides this, some early works~\cite{shi2021geometric, li2019gs3d, li2019stereo} predicted 2D box and depth information from camera inputs and lifted the results into 3D spaces via the geometric properties and constraints to obtain 3D prediction, gaining low generality limited by relying on feature engineering. 

Another branch implicitly utilizes depth information. The core of these methods~\cite{wang2021detr3d, liu2022petr, wang2023streampetr} lies in querying objects from image features, through the camera intrinsics/extrinsic matrix. The object queries in 3D space could project to the reference regions from camera inputs and aggregate the object presentation via transformer decoder. Apart from the previous methods, BEVFormer~\cite{li2022bevformer,li2022bevformerv2} project 2D features into the BEV plane via the BEV queries, which is aggregated from camera features similar to DETR3D~\cite{wang2021detr3d}.

\textbf{Depth Estimation} is a classical computer vision task. The previous general depth estimation methods mainly consist of single- and multi-view methods. The former~\cite{bhat2021adabins,zhou2020pattern,eigen2014depth, fu2018depth, ranftl2021vision} regard the depth estimation task as a regression or classification problem of depth distribution. These works adopt an encoder-decoder framework which is supervised by the given dense depth map. The latter~\cite{zhu2021deep, wei2021aa, wei2022bidirectional} usually construct a cost volume to regress disparities based on photometric consistency for effective depth prediction.
For camera-based 3D detection, previous works~\cite{li2023bevdepth, li2023bevstereo, reading2021categorical} introduce an extra depth estimation module to predict depth for accurate 3D object localization. CaDDN~\cite{reading2021categorical} learns a categorical depth distribution and then projects image features into BEV space. BEVDepth~\cite{li2023bevdepth} supervises depth learning with sparse LiDAR data and proposes a depth correction module to reduce depth error.

\begin{figure*}[thpb]
  \centering
   \includegraphics[width=0.99\textwidth]{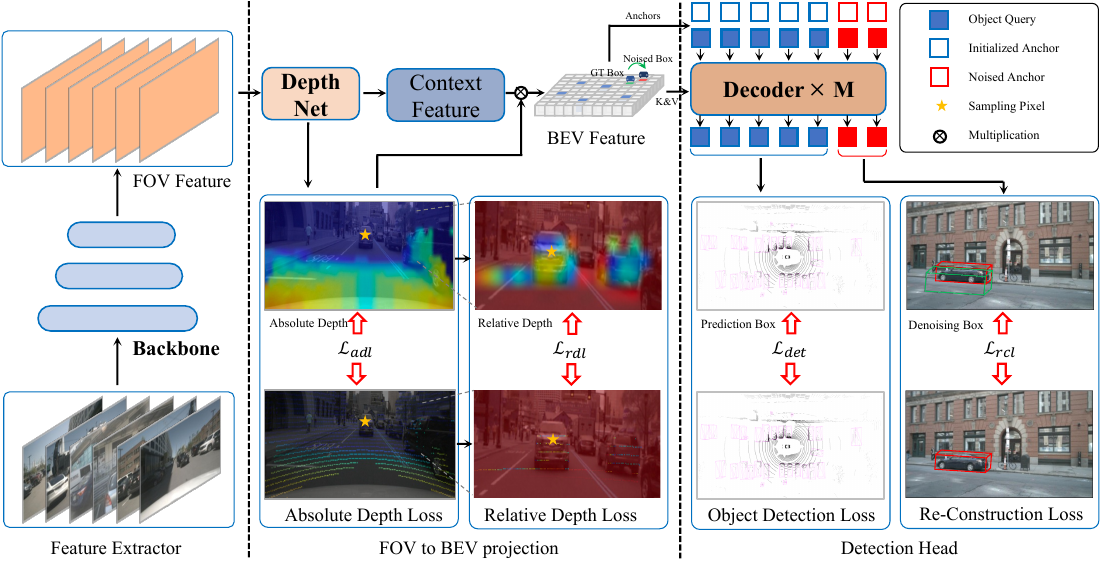}
  \vspace{-3mm}
  \caption{The overall architecture of our proposed detector consists of three parts: feature extractor, FOV to BEV translation, and detection head. For a given multi-camera input, we extract 2D features via a shared encoder and project them to 3D spaces via predicted depth, and the generated BEV features are fed into the detection head for object localization and recognition. The $L_{adl}$ and $L_{det}$ indicate the absolute depth loss and detection loss while $L_{rdl}$ and $L_{rcl}$ are relative depth loss and reconstruction loss in our proposed DE and DC scheme. Best view in color and more details are referred in Sec.~\ref{sec:method}.}
  \label{fig:ppl}
  \vspace{-5mm}
\end{figure*}

\textbf{Denoising Mechanism} aims to reconstruct the noised anchors to the ground truth to integrate the recognition calibration to the transformer head~\cite{li2022dndetr, zhang2022dino, yang2023neural}. which is first proposed~\cite{li2022dndetr} to accelerate the convergence of DETR-based detectors by introducing noised positional anchors and reconstructing them to the original box. Furthermore, DINO~\cite{zhang2022dino} introduces a contrastive denoising training strategy and gains promising improvements on the 2D object detection benchmark.  
In this paper, we introduce depth calibration, which adopts depth denoising to adjust the 3D object localization perturbation along depth axis for accurate 3D detection.

\section{Method}
\label{sec:method}

\subsection{The Overall Framework}
\label{sec:arc}
As shown in Fig.~\ref{fig:ppl}, we build up a baseline framework consisting of both the DE and DC schemes. It can be briefly divided into three parts.
The first is a shared encoder to extract the feature maps of each image. Given multi-camera inputs $\textbf{X}=\{\textbf{X}_1,\textbf{X}_2,...,\textbf{X}_N\}$, 
% where $N$ is the views number, 
where $N$ indicates the number of camera views.
we simply adopt ResNet followed FPN as the backbone to extract features $\textbf{F}=\{\textbf{F}_n\in \mathbb{R}^{C \times H \times W}\}_{n=1}^N$, where $H, W, C$ denote the height, the width, and the channel dimension.

The second part aims to transfer the multi-camera features from 2D to 3D space, \textit{i.e.} from Front-of-View (FOV) to Bird's-Eye-View (BEV).
Following BEVDepth~\cite{li2023bevdepth}, we first employ DepthNet to predict the depth distribution $\textbf{F}^d = \{\textbf{F}^d_n\in \mathbb{R}^{C_d \times H \times W}\}_{n=1}^N$ and context features $\textbf{F}^c = \{\textbf{F}^c_n\in \mathbb{R}^{C_c \times H \times W}\}_{n=1}^N$ from multi-view image features.
% which are the depth prediction and contextual information extracted from image features.
Here $C_d=118$ indicates the probabilities of the pixel-wise depth from $1$ to $118$ meters and $C_c=80$ is the contextual channel. Then we lift the multi-view 2D context features into 3D spaces via the prediction depth and camera intrinsics, and obtain one BEV features $\textbf{F}^b\in \mathbb{R}^{C_b \times H \times W}$ the same as BEVDepth. More details could be referred in BEVDepth~\cite{li2023bevdepth}.
To obtain the accurate depth, we supervise the depth estimation via the proposed DE scheme introduced in Sec.~\ref{sec:cdl}.

Finally, we adopt a simple transformer decoder-based head to localize the 3D objects from the BEV feature. Specifically, we randomly initialize object query content $\textbf{Q}_c\in \mathbb{R}^{m\times C}$ and reference learnable anchors $\textbf{Q}_r\in \mathbb{R}^{m\times 6}$, to aggregate the object presentations from BEV features in the specific reference region via a simple sequential deformable decoder. Here $m=900$ indicates the query number and $C=256$ denotes the query dimension.
Each reference anchor is represented as $6$ dimensions, which denotes the anchor location~($x,y,z$) and scale~($w,l,h$). The updated object queries are further fed into a detection head to predict the box location and a classification head to predict the class label.

Even if the DE scheme effectively improves the accuracy of depth prediction, prediction errors are still inevitable. This results in disturbances in the object localization along the depth axis. To address this, we further introduced the DC scheme to allow the model to adaptively mitigate such errors in Sec.~\ref{sec:col}.
Specifically, we simulate the depth prediction error and generate the extra object query with noised reference anchors $\textbf{R}'_r \in \mathbb{R}^{m'\times 6}$, as well as the corresponding query content $\textbf{Q}'_c\in \mathbb{R}^{m'\times C}$ associated with the object class label. 
By reconstructing the original object boxes during the training phase, we can implicitly integrate depth calibration capabilities into the detection model the same as~\cite{li2022dndetr}.

\subsection{Depth Estimation Scheme.}
\label{sec:cdl}

\begin{figure*}[thpb]
  \centering
  \includegraphics[width=0.99\textwidth]{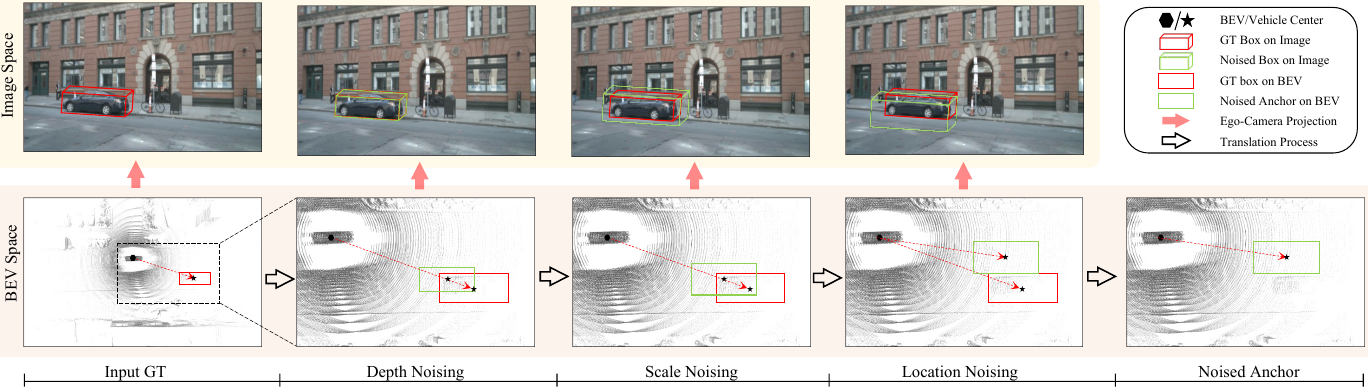}
  \caption{Noised anchors generation process. Given a ground truth box in the ego coordinate system, we add depth, scale, and location noise to generate the noised reference anchors, which indicate $f_d$,$f_s$, and $f_l$ in Eqn.~\ref{eq:6} respectively. We visualize the corresponding noised boxes in BEV and image space.}
  \label{fig:col}
  \vspace{-5mm}
\end{figure*}

In addition to directly using absolute depth as supervisory information, we also introduce relative depth to describe the depth relationship between two positions in the scene and use it as auxiliary supervision for the depth estimation module. For simplicity, we omit the view index $n$ and process all image views equally.
Given a prediction depth probability distribution $\textbf{F}^d\in \mathbb{R}^{C_d \times H \times W}$ of a specific image, we first adopt the corresponding LiDAR data and project to image space to obtain the sparse depth map $\textbf{G}^d \in \mathbb{R}^{C_d \times H \times W}$ as well as a mask $\textbf{M}\in \mathbb{R}^{H \times W}$. 
Here, $C_d=118$ and $\textbf{F}^d_i[c]\in (0,1)$ indicates the probability of the $i$-pixel belong to the $c$ meters depth, where $0\leq i < H\times W$ and $1\leq c\leq C_d$.
$\textbf{G}^d_{i}$ is a one-hot vector that records the real depth value of the $i$-pixel. $\textbf{M}$ denotes a mask and $\textbf{M}_{i}=0$ means no LiDAR point projects to corresponding position, otherwise $\textbf{M}_{i}=1$.

Following BEVDet~\cite{huang2021bevdet}, we first introduce the absolute depth loss and consider the depth estimation as a pixel-wise classification task, supervised by the cross-entropy loss.
\begin{align}
    \mathcal{L}_{adl} &= - \frac{1}{|\textbf{M}|} \sum_{i=0}^{H\times W} \sum_{c=1}^{C_d} \textbf{M}_{i} \cdot \textbf{G}^d_{i}[c] \cdot \text{log}(\textbf{F}^d_{i}[c]),
\end{align}
where $|\textbf{M}|$ is the summation of $\textbf{M}$.

Furthermore, we introduce relative depth as the auxiliary supervision to optimize the depth estimator, maintaining the structural context of the scene. To achieve this, we calculate the depth value from the $\textbf{F}^d$ and $\textbf{G}^d$ :
\begin{equation}
\begin{split}
    \widetilde{\textbf{F}}^d_{i} &= \sum_{c=1}^{C_d} c \cdot \textbf{F}^d_{i}[c],\\
    \widetilde{\textbf{G}}^d_{i} &= \text{argmax}(\textbf{G}^d_{i}),
\end{split}
\end{equation}
Here, we calculate the depth value in an integral manner, and the $\widetilde{\textbf{F}}^d_{i}$ and $\widetilde{\textbf{G}}^d_{i}$ denotes the predicted and the ground truth depth values of $i$-th pixels. 
Then the relative depth can be formulated as,
\begin{align}
\label{eq:cdl_od1}
    \textbf{R}_{jk} &= \widetilde{\textbf{F}}^d_j - \widetilde{\textbf{F}}^d_k, 
\end{align}
where $\textbf{R} \in \mathbb{R}^{HW\times HW}$ and $\textbf{R}_{jk}$ indicates the distance within $j$-th and $k$-th pixels along the depth axis.
Considering that a specific pixel might have significant relative depth differences with multiple pixels, which may result in numerical instability during the optimization, we introduce a normalization strategy to keep the values of relative depth within a certain range,
\begin{align}
\label{eq:cdl_od2}
    \textbf{R}_{jk} &= \frac{e^{-|\textbf{R}_{jk}|/\tau}}{\sum_{k=0}^{HW} e^{-|\textbf{R}_{jk}|/\tau}},
\end{align}
where $\tau$ is the temperature factor. According to the Eqn.~\ref{eq:cdl_od2}, the pixel pair within a shorter distance has a higher response.
We display the absolute depth as well as relative depth~(between the specific pixel in \textcolor{orange}{orange} star and other all pixels) in Fig.~\ref{fig:ppl}.
After achieving the normalization values from the depth ground truth, we can guide the learning of such relative depth using a Kullback–Leibler divergence loss.
\begin{align}
\label{eq:loss_rdl}
    % \mathcal{L}_{cdl} &= \textbf{R}^g * log \frac{\textbf{R}^g}{\textbf{R}^p}
    \mathcal{L}_{rdl} &= \text{Avg}(\textbf{R}^g \odot \text{log}(\textbf{R}^g/\textbf{R}^p)),
\end{align}
where $\textbf{R}^p$ and $\textbf{R}^g$ indicate the relative depth from the prediction and ground truth, $\odot$ is element-wise multiplication and Avg($\cdot$) indicates the average value of a matrix.

In practice, considering the memory efficiency and the depth relationships occur in a limited range, we only calculate the relative depth in local patches with the size $p\times p$, which are selected from the whole depth prediction map by using a sliding window, such that the relative depth scale is $p^2 \times p^2$ compared with $HW \times HW$, and the relative depth loss is the average value of Eqn.~\ref{eq:loss_rdl} for all local patches.

\subsection{Depth Calibration scheme.} 
\label{sec:col}
The proposed DC scheme aims to simulate the depth prediction errors and generate noised anchors, by reconstructing the extra noised queries to the real object location, the depth calibration capability could be implicitly embedded into the detection transformer.
In this section, we introduce the noised reference anchor generation process as shown in Fig.~\ref{fig:col} for the DC scheme.
Specifically, the process mainly disturbs the ground box in three aspects to generate noised reference anchors, including depth, scale, and location. Given a ground truth box in ego coordinate system $\mathcal{B} = (x,y,z,w,l,h)$, the noised box in Fig.~\ref{fig:col} could be formulated as:
\begin{align}
\label{eq:6}
    % \mathcal{B}' &= f_l(f_s(f_d(\mathcal{B}, \sigma_d), \sigma_s), \sigma_l)
    \mathcal{B}' &= f_l \oast f_s \oast f_d(\mathcal{B}; \sigma_d, \sigma_s, \sigma_l),
\end{align}
where  $\oast$ means assembling translation function $f_d, f_s, f_l$, which adds depth, scale, and location noise by factor $\sigma_d, \sigma_s, \sigma_l$. $\mathcal{B}'=(x',y',z',w',l',h')$ is the noised reference anchor. We describe the translation function as follows.

First, since the depth prediction occurs in 2D images, the newly added depth noise will simultaneously affect the anchor's location and scale in 3D space, which can be formulated as:
\begin{align}
    f_d(\mathcal{B}, \sigma_d) &= \sigma_d \cdot \begin{bmatrix} x,y,z,w,l,h \end{bmatrix}^T,
\end{align}
To prove this, we first consider projecting any points ($x,y,z$) to the 2D space with the depth information ($u,v,d$) via the translation parameter $\textbf{P}$:
\begin{align}
    d \begin{bmatrix} u, v, 1 \end{bmatrix}^T &= \textbf{P}\begin{bmatrix} x, y, z \end{bmatrix}^{T},
\end{align}
where ($u,v$) is the pixel position the 2D space and $d$ denotes the corresponding depth value of pixel ($u,v$). Adding depth noise with factor $\sigma_d$ will change the depth $d'=\sigma_d * d$, and the corresponding 3D points with depth noise is:
\begin{equation}
\begin{split}
    \begin{bmatrix} x', y', z' \end{bmatrix}^T &= \textbf{P}^{-1} d' \begin{bmatrix} u, v, 1 \end{bmatrix}^T = \sigma_d \cdot \begin{bmatrix} x, y, z \end{bmatrix}^T, 
\end{split}
\end{equation}
Based on the above translation, we conclude that adding depth noising will lead to constant changes in the point location. So, the box scale, which is the distances of two corner points will constantly change with the same factor, which can be presented as follows:
\begin{equation}
\begin{split}
    \begin{bmatrix} w', l', h' \end{bmatrix}^T &= \sigma_d \cdot \begin{bmatrix} w, l, h \end{bmatrix}^T, 
\end{split}
\end{equation}
Next, we add scale noise and location noise, which simply modified the object scale and location as follows,
\begin{equation}
\begin{split}
    f_s(\mathcal{B}, \sigma_s) &= \begin{bmatrix} x,y,z,\sigma_s w,\sigma_s l,\sigma_s h \end{bmatrix}^T, \\
    f_l(\mathcal{B}, \sigma_l) &= \begin{bmatrix} \sigma_l x,\sigma_l y,\sigma_l z,w,l,h \end{bmatrix}^T, 
\end{split}
\end{equation}
To increase the diversity of noised anchors, the noised factors are random values in a specific range controlled by hyper-parameters, \textit{e.g.,} $\sigma_d = Rand(1-\delta_d, 1+ \delta_d)$. Here we set $\delta_d = 0.5$, $ \delta_s, \delta_l=0.1$

Finally, as described in Sec.~\ref{sec:arc}, we produce noised reference anchors $\textbf{Q}'_r$ via the above formulation and feed extra noised queries to the shared detection head. We constrain the reconstruction loss with the following formulation:
\begin{align}
    \mathcal{L}_{rcl} = \mathcal{L}_{det}(\mathcal{H}(\textbf{Q}'_c, \textbf{Q}'_r)),
\end{align}
where $\mathcal{H}$ is the decoder head and $\mathcal{L}_{det}$ is the detection loss that includes the classification cross-entropy loss and regression L1 loss.

\subsection{Supervision.}
\label{sec:sup}
Following the aforementioned architecture in Sec.~\ref{sec:arc}, as well as the proposed depth estimation scheme in Sec.~\ref{sec:cdl} and depth calibration in Sec.~\ref{sec:col}, the supervision in our proposed including the two basic supervision such as detection loss and absolute depth loss, as well as two auxiliary supervision including relative depth loss and reconstruction loss:
\begin{align}
    \mathcal{L} = \mathcal{L}_{adl} + \mathcal{L}_{det}(\mathcal{H}(\textbf{Q}_c, \textbf{Q}_r)) + \alpha \mathcal{L}_{rdl} + \beta \mathcal{L}_{rcl}
\end{align}
where $\alpha=0.1$ and $\beta=1.0$ are manually set balance factors.

\section{EXPERIMENTS}

\subsection{Experiment settings.}
\label{sec:exp_setting}
\textbf{Dataset and Metric.} We conduct experiments based on NuScenes~\cite{caesar2020nuscenes} benchmarks following previous works~\cite{liu2022petr, li2022bevformer, huang2021bevdet, li2023dfa3d}. NuScenes is a large-scale autonomous driving dataset, which contains $700/150/150$ sequences for training, testing, and validation. We follow previous works to take the official evaluation metrics, including mean average precision (mAP), ATE, ASE, AOE, AVE, and AAE for our evaluations. These six metrics evaluate results from the center distance, translation, scale, orientation, velocity, and attribute. In addition, a comprehensive metric called NuScenes Detection Score (NDS) is also provided.

\textbf{Implementation Details.} 
For the detection head, we adopt the Transformer with $6$ deformable encoder layers and $6$ deformable decoder layers with $900$ queries. The detector is trained with $20$ epochs using AdamW optimizer with a base learning rate of $2\times 10^{-4}$ on 8 NVIDIA Tesla A100 GPUs.
To demonstrate the effectiveness, we apply the DE to multiple open-sourced methods that leverage depth prediction in their frameworks, such as BEVDepth~\cite{li2023bevdepth} and BEVStereo~\cite{li2023bevstereo}. 
We also realize DC in various transformer-based detectors, such as DETR3D~\cite{wang2021detr3d}, BEVFormer~\cite{li2022bevformer}, and PETR~\cite{liu2022petr}. All the models in ablation studies adopt ResNet50~\cite{he2016resnet} as the backbone for higher efficiency. To compare with previous state-of-the-art detectors, we adopt Swin-Base~\cite{liu2021swin} as our backbone and improve the performance via CBGS~\cite{zhu2019class} data re-sampling strategy, Exponential Moving Average~(EMA)~\cite{tarvainen2017mean}, and sequential long-term frames following BEVStrereo~\cite{li2023bevstereo}.
\begin{table}[!tbp]
\centering
\caption{Investigation of the DE scheme with different settings. RD, SW, KL indicate Relative Depth, sliding window, and KL loss. $p$ and $\tau$ indicate patch size and temperature factor.}
\vspace{-3mm}
\begin{tabu}{lcc} 
\tabucline[0.6pt]{-}
Method  & NDS$\uparrow$ & mAP$\uparrow$  \\
\hline
Baseline & 0.495  & 0.392  \\
Baseline+RD & 0.501 & 0.400  \\
Baseline+RD+SW($p=5$) & 0.503 & 0.401   \\
Baseline+RD+SW($p=7$) & 0.500 & 0.398   \\
Baseline+RD$^\dag$+SW($p=5$)+KL($\tau=4$)& 0.505 & 0.403   \\
Baseline+RD$^\dag$+SW($p=5$)+KL($\tau=8$)& \textbf{0.510} & \textbf{0.409}   \\
Baseline+RD$^\dag$+SW($p=5$)+KL($\tau=16$)& 0.508 & 0.406   \\
\tabucline[0.6pt]{-}
\end{tabu}
\label{tab:ab_cdl1}
\end{table}

\begin{table}[!tbp]
\centering
\caption{NDS and mAP on NuScene val set. $\alpha$ indicates balance factor.}
\vspace{-3mm}
\begin{tabu}{p{4.8cm}cc} 
\tabucline[0.6pt]{-}
Method  & NDS $\uparrow$ & mAP $\uparrow$  \\
\hline
Baseline & 0.495  & 0.392  \\
Baseline+DE($\alpha=0.01$) & 0.502 & 0.403  \\
Baseline+DE($\alpha=0.1$) & \textbf{0.510 }& \textbf{0.409}  \\
Baseline+DE($\alpha=0.5$) & 0.506 & 0.404  \\
\tabucline[0.6pt]{-}
\end{tabu}
\label{tab:ab_cdl2}
\end{table}

\begin{table}[!tbp]
\caption{Generalization Ability of DE scheme on multiple detectors.}
\vspace{-3mm}
\centering
\resizebox{\linewidth}{!}{
\begin{tabu}{c|cc|ccccc} 
\tabucline[0.6pt]{-}
Method  & NDS$\uparrow$ & mAP$\uparrow$ & mATE$\downarrow$ & mASE$\downarrow$ & mAOE$\downarrow$ & mAVE$\downarrow$ & mAAE$\downarrow$ \\
\hline
BEVDepth & 0.484 & 0.363  & 0.612 & \textbf{0.278} & 0.482 & 0.402 & 0.202 \\
BEVDepth+DE  & \textbf{0.493} & \textbf{0.370} & \textbf{0.600} & 0.280 & \textbf{0.477} & \textbf{0.371} & \textbf{0.189} \\
\hline
BEVStereo & 0.497 & 0.381 & 0.584 & 0.281 & \textbf{0.473} & 0.385 & 0.207 \\
BEVStereo+DE & \textbf{0.507} & \textbf{0.393} & \textbf{0.576} & \textbf{0.275} & 0.476 & \textbf{0.374} & \textbf{0.190} \\
\tabucline[0.6pt]{-}
Baseline & 0.495 & 0.392 & 0.627 & 0.327 & \textbf{0.460} & 0.396 & 0.204 \\
Baseline+DE & \textbf{0.510} & \textbf{0.409} & \textbf{0.615} & \textbf{0.298} & 0.471 & \textbf{0.374} & \textbf{0.193} \\
\tabucline[0.6pt]{-}
\end{tabu}}
\label{tab:cdl}
\end{table}

\begin{table*}[!tbp]
\centering
\resizebox{\linewidth}{!}{
\begin{tabu}{c|c|c|c|cc|ccccc} 
\tabucline[0.8pt]{-}
Method & Publication & Backbone & LiDAR & NDS $\uparrow$ & mAP $\uparrow$ & mATE $\downarrow$ & mASE $\downarrow$ & mAOE $\downarrow$ & mAVE $\downarrow$ & mAAE $\downarrow$ \\
\hline
BEVDet~\cite{huang2021bevdet} & Arxiv & R50 & \ding{55} & 0.379 & 0.298 & 0.725 & 0.279 & 0.589 & 0.860 & 0.245 \\
DETR3D~\cite{wang2021detr3d} & CoRL2021 & R50 & \ding{55} & 0.373 & 0.302 & 0.811 & 0.282 & 0.493 & 0.979 & 0.212 \\
PETR~\cite{liu2022petr} & ECCV2022 & R50 & \ding{55} & 0.403 & 0.339 & 0.748 & 0.273 & 0.539 & 0.907 & 0.203 \\
BEVFormer~\cite{li2022bevformer} & ECCV2022 & R50 & \ding{55} & 0.354 & 0.252 & 0.900 & 0.294 & 0.655 & 0.657 & 0.216 \\
PETRv2~\cite{liu2022petrv2} & ICCV2023 & R50 & \ding{55} & 0.494 & \underline{0.398} & 0.690 & 0.273 & 0.467 & 0.424 & 0.195 \\
BEVDet4D~\cite{huang2022bevdet4d} & Arxiv & Swin-T & \ding{51} & 0.476 & 0.338 & 0.672 & 0.274 & \underline{0.460} & \textbf{0.337} & \textbf{0.185} \\
BEVDepth~\cite{li2023bevdepth} & AAAI2023 & R50 & \ding{51} & 0.478 & 0.359 & 0.612	& \underline{0.269} & 0.507 & 0.409 & 0.201 \\
BEVStereo~\cite{li2023bevstereo} & AAAI2023 & R50 & \ding{51} & \underline{0.500} & 0.372 & \textbf{0.598} & 0.270 & \textbf{0.438} & \underline{0.367} & \underline{0.190} \\
Ours & ICRA2024 & R50 & \ding{51} & \textbf{0.517} & \textbf{0.419} & \underline{0.611} & \textbf{0.266} & 0.465 & 0.368 & 0.211 \\
\hline
DETR3D$^\ddag$~\cite{wang2021detr3d} & CoRL2021 & R101 & \ding{55} & 0.434 & 0.349 & 0.716 & 0.268 & 0.379 & 0.842 & 0.200 \\
PETR$^\ddag$~\cite{liu2022petr} & ECCV2022 & R101 & \ding{55} & 0.442 & 0.370 & 0.711 & 0.267 & 0.383 & 0.865 & 0.201 \\
BEVFormer~\cite{li2022bevformer} & ECCV2022 & R101-DCN & \ding{55} & 0.517 & 0.416 & 0.672& 0.274 & 0.369& 0.397& 0.191 \\
PETRv2$^\ddag$~\cite{liu2022petrv2} & ICCV2023 & R101 & \ding{55} & 0.524 & 0.421 & 0.681 & 0.267 & 0.357 & 0.377 & \underline{0.186} \\
PolarFormer~\cite{jiang2023polarformer} & AAAI2023 & R101-DCN & \ding{55} & 0.528 & \underline{0.432} & 0.648 & 0.270 & \underline{0.348} & 0.409 & 0.201 \\
BEVDet4D~\cite{huang2022bevdet4d} & Arxiv & Swin-B & \ding{51} & \underline{0.545} & 0.421 & 0.579 & \textbf{0.258} & \textbf{0.329} & \textbf{0.301} & 0.191 \\
FCOS3D~\cite{wang2021fcos3d} & ICCV2021 & R101 & \ding{51} & 0.415 & 0.343 & 0.725 & 0.263 & 0.422 & 1.292 & \textbf{0.153} \\
BEVDepth~\cite{li2023bevdepth} & AAAI2023 &R101 & \ding{51} & 0.535 & 0.412 & \underline{0.565} & 0.266 & 0.358 & \underline{0.331} & 0.190 \\
Ours & ICRA2024 & Swin-B & \ding{51} & \textbf{0.559} & \textbf{0.457} & \textbf{0.543} & \underline{0.259} & 0.354 & 0.349 & 0.193 \\
\tabucline[0.8pt]{-}
\end{tabu}}
\caption{Performances comparison on NuScene val set. $^\ddag$ indicates using pre-trained model from FCOS3D. LiDAR indicates supervising the depth estimation by ground truth generated from LiDAR data. The best and second are in \textbf{bold} and \underline{underlined}.}
\label{tab:overall}
\vspace{-0.6cm}
\end{table*}

\subsection{Ablation Study for Depth Estimation Scheme.} 
\label{sec:exp_ab_cdl}

To evaluate the depth estimation scheme, we conduct experiments with several settings as shown in Tab.~\ref{tab:ab_cdl1}. Here, the RD indicates using the relative depth in Eqn.~\ref{eq:cdl_od1} to define the relationships and supervised by L1 loss in contrastive to KL loss, the SW indicates select relative regions in a slide window manner with window side $p$, in contrast to global manner, RD$^\dag$ and KL means using depth relations in Eqn.~\ref{eq:cdl_od2} with temperature factor $\tau$ and supervised by the Kullback–Leibler divergence loss. 
With our proposed DE scheme, the best setting yields results with $51.0\%/40.9\%$ in terms of NDS and mAP, exceeding the baseline model $49.5\%/39.2\%$, and gains $+1.5\%/+1.7\%$ improvements. Besides, we conduct experiments of the hyper-parameters for balance factor $\alpha$ in Tab.~\ref{tab:ab_cdl2}. As an auxiliary supervision, the $\alpha=0.1$ could gain the best improvements.

To demonstrate the broad applicability of the proposed DE, we apply the DE scheme to multiple open-sourced methods that adopt the depth estimator.
As reported in Tab.~\ref{tab:cdl}, DE brings consistent improvements and demonstrates its capability across different models. 
Without bells and whistles, the DE scheme improves BEVDepth and BEVStereo for $+0.9\%/+0.7\%$ and $+1.0\%/+1.2\%$ NDS/mAP improvements without introducing any extra inference cost.

\begin{table}[!tbp]
\centering
\caption{Investigation of the DC scheme with different settings. DN, LN, and SN indicate Depth, Location, and Scale Noise.}
\vspace{-3mm}
\resizebox{\linewidth}{!}{
\begin{tabu}{lcc} 
\tabucline[0.6pt]{-}
Method  & NDS$\uparrow$ & mAP$\uparrow$  \\
\hline
Baseline & 0.495  & 0.392  \\
Baseline+DN($\delta_d=0.1$) & 0.499 & 0.400  \\
Baseline+DN($\delta_d=0.5$) & 0.504 & 0.403  \\
Baseline+DN($\delta_d=0.5$)+LN($\delta_l=0.1$) & 0.506 & 0.406   \\
Baseline+DN($\delta_d=0.5$)+LN($\delta_l=0.5$) & 0.506 & 0.402   \\
Baseline+DN($\delta_d=0.5$)+LN($\delta_l=0.1$)+SN($\delta_s=0.1$) & \textbf{0.511} & \textbf{0.412}   \\
Baseline+DN($\sigma_d=0.5$)+LN($\delta_l=0.1$)+SN($\delta_s=0.5$) & 0.508 & 0.407   \\
\tabucline[0.6pt]{-}
\end{tabu}}
\label{tab:ab_col1}
\end{table}

\begin{table}
\caption{Generalization Ability of DC scheme on multiple detectors.}
\vspace{-3mm}
\centering
\resizebox{\linewidth}{!}{
\begin{tabu}{c|cc|ccccc} 
\tabucline[0.6pt]{-}
Method  & NDS$\uparrow$ & mAP$\uparrow$ & mATE$\downarrow$ & mASE$\downarrow$ & mAOE$\downarrow$ & mAVE$\downarrow$ & mAAE$\downarrow$ \\
\hline
DETR3D & 0.373 & 0.302 & 0.811 & 0.282 & \textbf{0.493} & 0.979 & 0.212 \\
DETR3D+DC  & \textbf{0.391} & \textbf{0.323} & \textbf{0.799} & \textbf{0.259} & 0.495 & \textbf{0.952} & \textbf{0.203}  \\
\hline
BEVFormer & 0.354 & 0.252 & 0.900 & \textbf{0.294} & 0.655 & 0.657 & \textbf{0.216} \\
BEVFormer+DC & \textbf{0.380} & \textbf{0.276} & \textbf{0.880} & 0.295 & \textbf{0.588} & \textbf{0.596} & 0.221 \\
\hline
PETR & 0.403 & 0.339 & 0.748 & 0.273 & 0.539 & 0.907 & 0.203 \\
PETR+DC & \textbf{0.420} & \textbf{0.358} &  \textbf{0.725} & \textbf{0.269} & \textbf{0.514} & \textbf{0.887} & \textbf{0.192} \\
\hline
Baseline & 0.495 & 0.392 & 0.627 & 0.327 & 0.460 & 0.396 & \textbf{0.204} \\
Baseline+DC & \textbf{0.511} & \textbf{0.412} & \textbf{0.601} & \textbf{0.318} & \textbf{0.443} & \textbf{0.371} & 0.215 \\
\tabucline[0.6pt]{-}
\end{tabu}}
\label{tab:col2}
\end{table}

\subsection{Ablation Study for Depth Calibration Scheme.}
\label{sec:exp_ab_col}

As shown in Tab.~\ref{tab:ab_col1}, we conduct experiments with several settings and hyper-parameters to analyze the depth calibration scheme. 
Here, the DN, LN, and SN indicate the depth noising, location noising, and scale noising respectively. From the results, we can observe that adding the depth noising could gain significant improvements by $+0.9\%/+1.1\%$ in terms of NDS and mAP. Furthermore, denoising the location and scale errors could continuously boost the performances and yield $+1.6\%/+2.0\%$ improvements.

To demonstrate the broad applicability of the proposed DC, we also apply DC to multiple open-sourced methods that employ transformer-based detection heads.
As illustrated in Tab.~\ref{tab:col2}, the DC scheme consistently enhances various methods. Specifically, the DE scheme boosts DETR3D, BEVFormer, and PETR by $+1.8\%/+2.1\%$, $+2.6\%/+2.4\%$, and $+1.7\%/+1.9\%$  NDS/mAP.

\subsection{Ablation Study for Module Pruning.} 
\label{sec:aba_prune}
In Tab.~\ref{tab:ab_cdl1} and Tab.~\ref{tab:ab_col1}, we analyze the impacts of hyper-parameters and settings in the proposed DE and DC schemes, as well as the performance improvements based on ResNet50 backbone over the baseline model. Furthermore, we conduct experiments to implement them in a cascade manner as shown in Tab.~\ref{tab:overall}. We can observe that the proposed detector gains $51.7\%/41.9\%$ NDS/mAP, and achieves $+2.2\%/+2.7\%$ improvements with ResNet50, which is higher than the module pruning version, demonstrating that the DE and DC could boost the performances from different aspects and jointly promote the detection results to a certain extent. 

\subsection{Comparison with the State-of-The-Art Methods.}
\label{sec:exp_sota}
We aim to prove the efficacy of our proposed approach by making a comparison with previous state-of-the-art methods
as shown as Tab.~\ref{tab:overall}. For a fair comparison, all the detectors adopt R50/Swin-T and R101/Swin-B backbones. 
From the results, we can observe that our proposal could gain the best NDS/mAP performance. Specifically, our proposal could gain $51.7\%/41.9\%$ NDS/mAP with ResNet50 backbone, and achieve $+1.7\%/+4.7\%$ improvements against BEVStereo. 
When enlarging the backbone to Swin-Base, our proposal gains $55.9\%/45.7\%$ NDS/mAP, and gains a significant margin over the previous SOTA methods, such as $+3.1\%/+2.5\%$ over PolarFormer and $+1.4\%/+3.6\%$ over BEVDet4D. The exciting performance and significant promotion demonstrate the effectiveness of our proposal.

\section{CONCLUSIONS}
In this paper, we introduce a cascade framework, aiming to improve the accuracy of camera-based 3D object detection by promoting depth prediction in both feature lifting and object localization phases.
Our method comprises two depth-aware learning paradigms named depth estimation and depth calibration. The depth estimation (DE) employs depth relationships as the constraints in the training phase to ensure precise depth estimation. Conversely, the depth calibration (DC) scheme integrates a depth denoising mechanism, equipping the detection transformer with depth calibration capabilities. Both the DE and DC  can be seamlessly integrated into the proposed detectors and previous cutting-edge detectors to significantly elevate detection performance without adding extra inference costs. 

\section*{ACKNOWLEDGMENT}
The work is partially supported by the Young Scientists Fund of the National Natural Science Foundation of China under grant No.62106154, by the Natural Science Foundation of Guangdong Province, China (General Program) under grant No.2022A1515011524, and by Shenzhen Science and Technology Program JCYJ20220818103001002 and ZDSYS20211021111415025, and by the Guangdong Provincial Key Laboratory of Big Data Computing, The Chinese University of Hong Kong (Shenzhen).

\end{document}